\documentclass[]{bytedance}

\usepackage{amsmath, amssymb}
\usepackage{makecell}
\usepackage{xspace}
\usepackage{fontawesome5}

\newcommand{\method}{SplitMoE\xspace}
\newcommand{\eg}{\textit{e.g.}}
\definecolor{scoreLow}{RGB}{255,255,255}
\newcommand{\lowscore}[1]{\cellcolor{scoreLow}#1}

\title{Breaking the Uniformity Trap:\\ Scaling Video Diffusion Model via SplitMoE}

\author{%
\parbox{\textwidth}{\centering
Yu Xu$^{*1,2}$, Yuxin Zhang$^{2\dagger}$, Xiao Yang$^{3}$, Haotian Yang$^{2}$, Yizhi Wang$^{2}$\\[1mm]
Xinwei Huang$^{2}$, Minxuan Lin$^{2}$, Angtian Wang$^{2}$, Chongyang Ma$^{2}$, Fan Tang$^{4}$
}}

\affiliation{%
\parbox{\textwidth}{\centering\small
$^1$University of Chinese Academy of Sciences,
$^2$ByteDance\\[1mm]
$^3$Canva Research,
$^4$University of Science and Technology Beijing
\vspace{-1mm}

}}

\contribution[\dagger]{Corresponding Author}

\let\templateabstract\abstract
\DeclareDocumentEnvironment{abstract}{+b}{\templateabstract{#1}\global\let\abstractlist\abstractlist}{}
\begin{abstract}
Mixture-of-Experts (MoE), popularized by large language models, is a promising paradigm for scaling visual generative models.
However, conventional token-wise MoE routes tokens independently within a homogeneous expert pool and regularizes expert usage toward uniformity, making it poorly matched to video data that is spatiotemporally redundant and semantically long-tailed.
We show that existing visual MoEs fall into a uniformity trap: semantically under-organized routing, compounded by uniform expert-usage regularization, scatters coherent patches across disparate experts, causing routing fragmentation and structural distortion.
To address this, we propose \method, a split-role sparse architecture that breaks the shackles of uniformity.
To accommodate the inherent semantic imbalance, we explicitly bifurcate the expert pool into semantic experts and generic experts, with semantic experts capturing high-level semantic abstraction and generic experts preserving residual visual information and flexible generative capacity.
Leveraging prototype-guided routing and pull-push regularization, \method enables tokens to cluster naturally by semantic attributes rather than arbitrary balancing constraints.
Extensive results show that under an equivalent activated-parameter budget, \method outperforms traditional load-balanced MoEs in convergence speed, routing coherence, and video generation quality across standard benchmarks.
By revealing an emergent coarse-to-fine denoising logic, \method provides the community with a modality-aware scaling path, serving as a critical reference for building large-scale video world models.
\end{abstract}

\date{Sept 30, 2026}
\checkdata[Venue]{NeurIPS 2026 (Spotlight)}
\checkdata[\faGlobe\ Project Page]{\url{https://yuci-gpt.github.io/SplitMoE/}}

\begin{document}
\maketitle
\renewcommand{\thefootnote}{\fnsymbol{footnote}}
\footnotetext[1]{Work done during an internship at ByteDance.}
\renewcommand{\thefootnote}{\arabic{footnote}}
\suppressfloats[t]

\begin{figure}[t]
    \centering
    \includegraphics[width=1\textwidth]{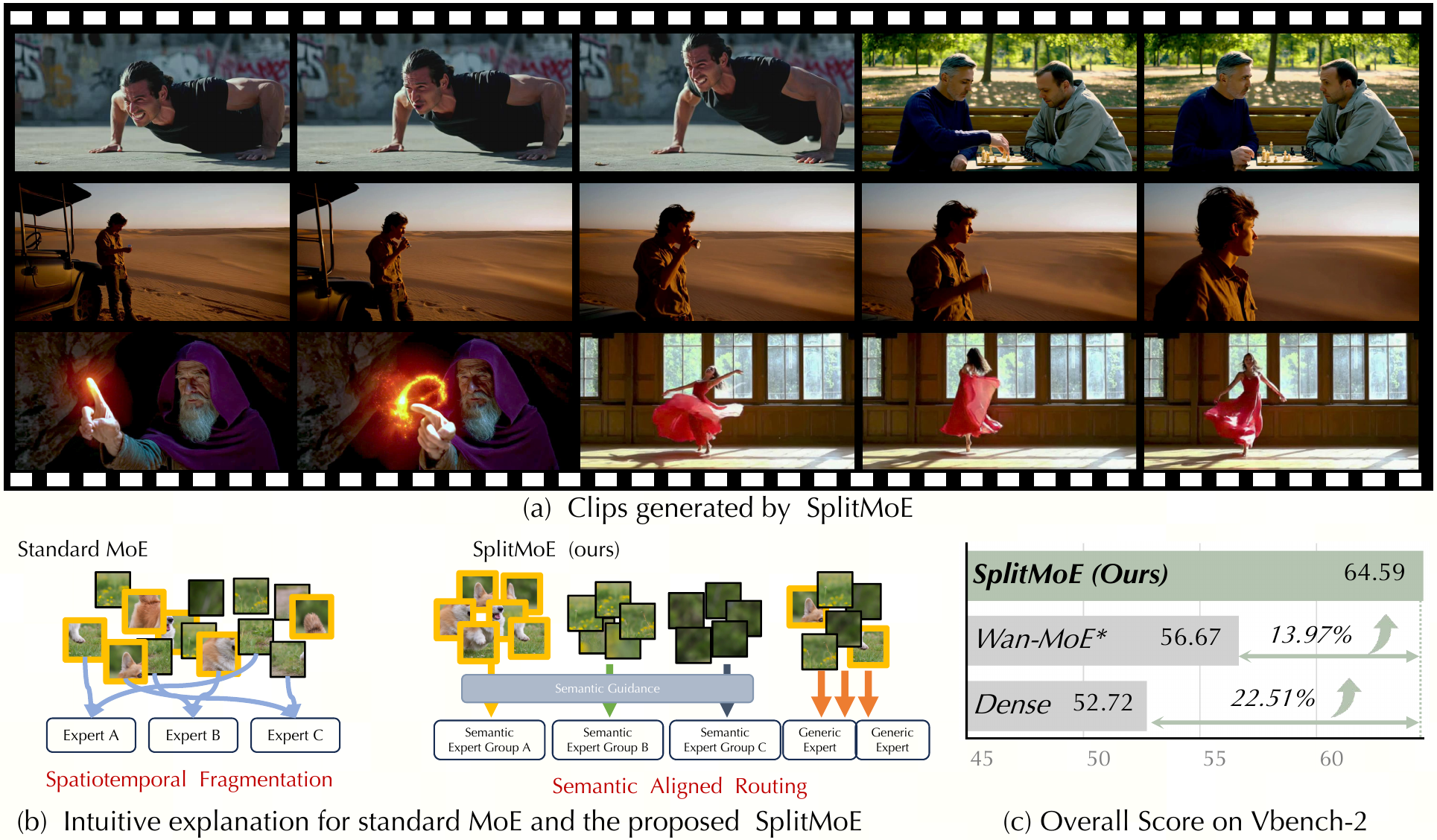}
    \vspace{-4mm}
    \caption{\method scales video diffusion models by introducing semantic-aligned expert partitioning, reducing spatiotemporal fragmentation, and generating higher-quality videos than dense fine-tuning and WAN with a standard MoE.
    }
    \label{fig:teaser}
\end{figure}

\section{Introduction}
\label{sec:intro}

Diffusion Transformers (DiTs)~\cite{peebles2023scalable} have made significant advances in high-fidelity video generation~\cite{openai2024sora, googledeepmind2024veo, gao2025seedance, kuaishou2024kling}, yet scaling their parameters to capture complex motion remains computationally prohibitive.
Mixture-of-Experts (MoE) provides a promising path toward scalable modeling by decoupling total model capacity from active FLOPs.
However, standard MoE architectures are developed for language models~\cite{lepikhin2020gshard, fedus2022switch, du2022glam}; nevertheless, current visual generative models~\cite{fei2024scaling, zheng2025dense2moe, shi2025diffmoe} often naively adopt these language-centric designs, taking their cross-modal effectiveness for granted and assuming that mechanisms optimized for discrete syntax will naturally generalize to the highly redundant and heterogeneous spatiotemporal tokens of video.

In this study, we revisit one of the fundamental settings of standard MoE: load-balancing mechanisms which enforce tokens' statistical uniformity across available experts when training.
We argue that such unique distributional characteristics are poorly aligned with the unstructured and imbalanced nature of video data.
While language tokens are often organized into distinguishable semantic units by discrete syntax, video tokens are highly imbalanced across different regions.
Unconstrained sparse routing in video diffusion features can therefore produce \textit{spatiotemporal fragmentation}: neighboring patches from the same coherent object may be dispatched to different experts, while visually redundant background patches may dominate multiple experts (as illustrated in Fig.~\ref{fig:teaser}(b)).

\begin{figure}[tb]
  \centering
  \includegraphics[width=\linewidth]{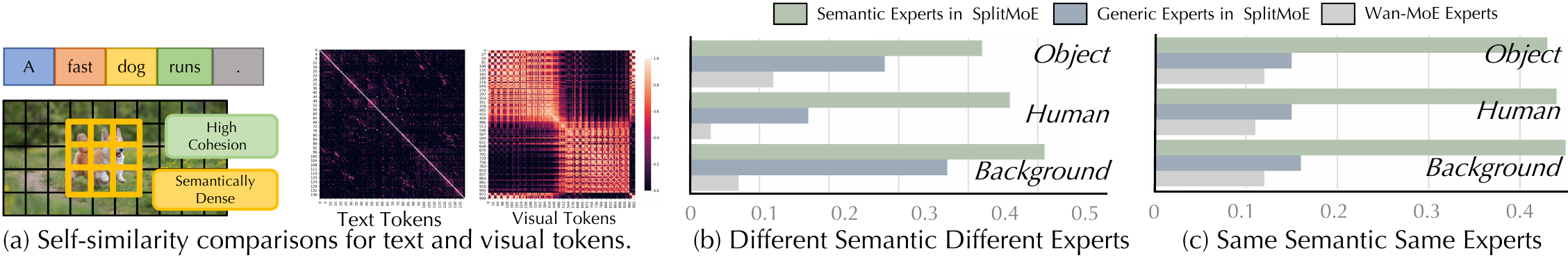}
  \vspace{-1.2em}
  \caption{
  Motivation for semantic-aware Video-MoE routing. Visual tokens exhibit much stronger cohesion than text tokens, making uniform MoE routing prone to fragmented assignments. Compared with standard MoE, \method routes tokens from different semantic regions to more distinguishable experts while assigning tokens within the same semantic region to more consistent expert groups.
  }
  \label{fig:intro_motivation}
\end{figure}

To investigate such a ``uniformity trap'', we compare the self-similarity patterns of language and visual tokens in Fig.~\ref{fig:intro_motivation}(a).
Unlike language tokens exhibiting relatively high entropy and low cohesion, visual tokens are densely correlated and spatially continuous.
Directly routing such tokens leads the experts to redundantly specialize in ubiquitous visual patterns rather than distinct semantic entities.
We further group video tokens with off-the-shelf semantic segmentation masks and analyze their routed experts in Fig.~\ref{fig:intro_motivation}.
Two complementary properties are measured: semantic distinctiveness (Fig.~\ref{fig:intro_motivation}(b))  evaluates whether different semantic classes are assigned to more category-specific experts, and intra-sample routing dominance (Fig.~\ref{fig:intro_motivation}(c)) evaluates whether tokens from the same semantic region are consistently routed to the same expert.
High values in both metrics indicate routing that is discriminative across semantic classes and coherent within each semantic region.
Standard MoE has consistently low semantic distinctiveness and routing dominance across most semantic classes, indicating that tokens from the same region are often dispersed across experts.
Enforcing strict uniform allocation may therefore amplify spatiotemporal fragmentation, since tokens within semantically coherent regions are forced into uniform routing patterns.

Motivated by these observations, we propose \method, a structured routing framework that separates semantic abstraction from general visual residual modeling. As shown in Fig.~\ref{fig:teaser}(b), \method partitions experts into a semantic branch and a generic branch.
The semantic branch is guided to route tokens from the same semantic region to consistent expert groups and to assign different semantic regions to more distinguishable expert groups, thereby improving semantic coherence while avoiding expert collapse. This effect is supported by the green bars in Fig.~\ref{fig:intro_motivation}: compared with standard MoE, our semantic branch achieves higher routing dominance across all semantic classes and stronger semantic distinctiveness for most categories.
This design addresses the fragmentation observed in common video MoE routing, where semantically similar regions may be split across unrelated experts.
As shown in Fig.~\ref{fig:teaser}(c), semantically coherent token assignment encourages experts to develop more specialized capabilities, leading to better generation performance.
Meanwhile, the generic branch retains flexible capacity for residual variations that are not well captured by discrete semantic grouping.
In this way, \method avoids forcing all visual tokens into a single uniform routing space.
To ensure meaningful specialization and prevent semantic collapse, we introduce a Prototype-Guided Semantic Routing mechanism. We use prototypes as a bridge between visual model features and DiT visual tokens, guiding the router toward semantic-aware expert assignment.

More importantly, we demonstrate that this role-aware decoupling aligns with the semantic-to-detail generation order of the diffusion process~\cite{zhang2023prospect} through Emergent Temporal Specialization. Without relying on any explicit timestep-conditioned routing constraints, \method naturally routes more capacity to Semantic Experts during the early, high-noise structural phase, and shifts more expert weights to Generic Experts during the final denoising stages. This emergent behavior validates the intuition behind our semantic-generic partition and establishes an effective paradigm for modeling heterogeneous video tokens.
In summary, our core contributions are threefold:
\begin{itemize}
    \item We propose \method, a sparse architecture tailored for video generation. By decoupling visual modeling into Semantic and Generic Experts, \method addresses the inherent cohesion of video tokens and mitigates spatiotemporal fragmentation and temporal flickering.

    \item We introduce Prototype-Guided Semantic Routing, which uses learnable prototypes to bridge reconstructive visual features and DiT visual tokens. It enables semantic-aware expert assignment, reduces expert homogenization, and promotes meaningful specialization while preserving flexible residual modeling.

    \item Under matched activated-parameter budgets, \textbf{\method} outperforms standard MoE and remains competitive with other baselines of comparable activated parameter counts. We provide empirical insights into Video-MoE routing dynamics and timestep analyses, which reveal a coarse-to-fine routing pattern consistent with denoising.
\end{itemize}

\section{Related Work}
\label{sec:related}

\textbf{Large-scale text-to-video generation.}
Text-to-video generation has advanced rapidly with the scaling of diffusion models and transformer architectures. Closed-source systems~\cite{runway2023gen2, openai2024worldsim, openai2025sora2, googledeepmind2024veo, kuaishou2024kling, lumalabs2024dreammachine, runway2024gen3, minimax2024hailuo, shengshu2024vidu} have set high standards in visual fidelity, physical plausibility, and long-video generation. Meanwhile, open-source models have shifted from 3D U-Nets~\cite{blattmann2023stable} to Diffusion Transformers~\cite{peebles2023scalable}, with recent systems~\cite{hong2022cogvideo, kong2024hunyuanvideo, peng2025open, team2025longcat, wan2025} combining DiT backbones, advanced autoencoders, and multimodal language models to approach proprietary performance. Despite this progress, open-source models still suffer from spatio-temporal semantic inconsistency, structural distortion, and temporal jitter, motivating more effective semantic representation and token-level control during generation.

\textbf{Mixture of experts for visual generation.}
MoE effectively scales LLMs~\cite{lepikhin2020gshard, fedus2022switch, du2022glam} by offering large capacity with manageable inference cost through sparse routing. This paradigm has extended to visual generation, where recent works~\cite{fei2024scaling, cao2025hunyuanimage, zheng2025dense2moe, shi2025diffmoe, xu2026tag} show that sparse architectures improve diffusion transformers for image synthesis. For video generation, existing methods~\cite{wan2025} adopt coarse timestep-level MoE routing, assigning one expert to all tokens at each denoising stage. While effective, this overlooks the spatio-temporal heterogeneity within video frames. Token-level routing can address this limitation but introduces another challenge: conventional load-balancing objectives~\cite{gale2023megablocks} uniformly scatter tokens across experts, disrupting strong video semantic correlations and temporal consistency.

Recent works have addressed homogeneous routing in static image generation. ProMoE~\cite{wei2025routing} uses prototype-guided routing to segregate conditional and unconditional tokens, while MammothModa2~\cite{shen2025mammothmoda2} decouples generation and understanding experts in a unified autoregressive-diffusion framework. However, they focus on conditioning types or task modalities rather than video structure. In contrast, \method targets the long-tailed spatio-temporal redundancy of video by splitting experts into semantic and generic pools, decoupling high-level structural abstraction from localized high-frequency reconstruction.

\section{Method}
\subsection{Preliminaries}

\begin{figure}%
  \centering
  \includegraphics[width=1\textwidth]{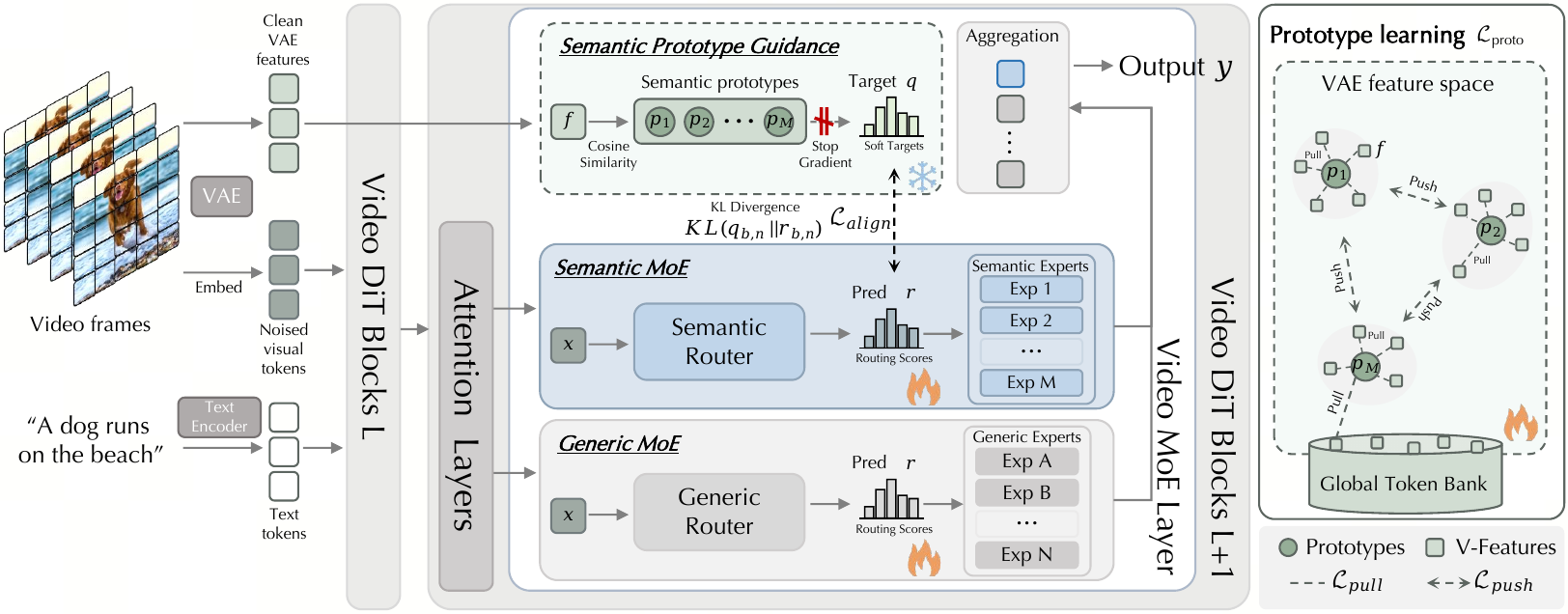}
  \vspace{-1.5em}
  \caption{Pipeline of \method. \method splits each Video MoE layer into Semantic and Generic MoE branches, where VAE-prototype induced soft targets guide semantic routing while the generic branch preserves flexible residual modeling capacity.
Learnable prototypes are optimized in the VAE feature space with pull-push regularization and a global token bank, encouraging semantic experts to form diverse and well-covered visual concepts.}
  \vspace{-1.5em}
  \label{fig:pipeline}
\end{figure}

A standard Mixture-of-Experts (MoE) layer~\cite{shazeer2017outrageously} replaces a dense feed-forward network with a set of experts $\mathcal{E}=\{E_1,\ldots,E_M\}$ and a router that activates only a small subset of experts for each token:
\begin{equation}
\mathbf{y} = \sum_{i\in\mathcal{I}(\mathbf{x})} g_i(\mathbf{x})E_i(\mathbf{x}), \qquad |\mathcal{I}(\mathbf{x})|=K\ll M,
\end{equation}
where $\mathcal{I}(\mathbf{x})$ denotes the set of selected experts and $g_i(\mathbf{x})$ represents the corresponding routing weight. Most existing MoE architectures originate from language modeling~\cite{fedus2022switch}, where load-balancing objectives are commonly employed. Specifically, an auxiliary loss is typically added during training to penalize routing skew, forcing a uniform distribution of tokens across all available experts to prevent expert collapse and maximize capacity utilization.

\subsection{\method: Split-Role Visual Experts}

We propose \method, which explicitly decouples the MoE layer into specialized functional roles. By utilizing continuous VAE features as a teacher signal to anchor tokens to learnable semantic prototypes, our approach separates high-level semantic abstraction from low-level generic reconstruction.

\paragraph{Decoupled expert partitioning.}
We partition the standard expert pool $\mathcal{E}$ into two disjoint subsets, specifically the semantic experts $\mathcal{E}^{\rm sem}$ and the generic experts $\mathcal{E}^{\rm gen}$, formulated as:
\begin{equation}
\mathcal{E} = \mathcal{E}^{\rm sem} \cup \mathcal{E}^{\rm gen}, \qquad \mathcal{E}^{\rm sem}\cap\mathcal{E}^{\rm gen}=\emptyset, \qquad M=M_s+M_g.
\end{equation}
Similar to an artist sketching a structural outline before rendering generic visual elements, semantic experts model high-level structures, while generic experts capture local appearance and generative residuals. This bifurcation alleviates capacity interference from forcing a single isotropic expert pool to jointly optimize low-frequency semantics and high-frequency generic features.

Given a patchified video token $\mathbf{x}\in\mathbb{R}^{C}$, the router initially projects the token to compute the semantic-generic affinities:
\begin{equation}
\mathbf{h}=\phi(\mathbf{W}_1\mathbf{x}), \qquad z_{e} = \alpha \left\langle \frac{\mathbf{h}}{\|\mathbf{h}\|_2}, \frac{\mathbf{w}_e}{\|\mathbf{w}_e\|_2} \right\rangle, \qquad s_{e}=\sigma(z_{e}),
\end{equation}
where $\phi$ denotes the GELU activation function and $\alpha$ represents a learnable, clipped scale that controls the sharpness of the sigmoid affinities. Unlike a global softmax operation that forces competition among all experts, the utilization of independent sigmoid scores $s_{e}$ allows a single token to be highly compatible with both a semantic expert and a generic expert simultaneously.

The logits and scores are divided into the respective semantic and generic groups, enabling the independent selection of the top $K_s$ and $K_g$ experts:
\begin{equation}
\mathcal{I}^{\rm sem} = \operatorname{TopK}(\tilde{\mathbf{s}}^{\rm sem},K_s), \qquad \mathcal{I}^{\rm gen} = \operatorname{TopK}(\tilde{\mathbf{s}}^{\rm gen},K_g), \qquad \mathcal{I} = \mathcal{I}^{\rm sem} \cup \mathcal{I}^{\rm gen}.
\end{equation}
Here, $\tilde{\mathbf{s}}$ represents the biased routing scores (detailed in Sec.~\ref{sec:loss_free}), which are utilized solely for the discrete expert selection step and may incorporate exploration noise or non-gradient balancing biases. However, to maintain routing fidelity, the final mixture weights are consistently computed in a dynamic manner from the clean raw scores:
\begin{equation}
g^{\rm sem}_{e} = \frac{s_{e}} {\sum_{j\in\mathcal{I}^{\rm sem}}s_{j}+\epsilon}, \qquad g^{\rm gen}_{e} = \frac{s_{e}} {\sum_{j\in\mathcal{I}^{\rm gen}}s_{j}+\epsilon}.
\end{equation}
The final MoE output aggregates the specialized contributions:
\begin{equation}
\mathbf{y} = \sum_{e\in\mathcal{I}^{\rm sem}} g^{\rm sem}_{e}E_e(\mathbf{x}) + \sum_{e\in\mathcal{I}^{\rm gen}} g^{\rm gen}_{e}E_e(\mathbf{x}).
\end{equation}
Crucially, a fixed active expert budget is maintained by ensuring that $K=K_s+K_g$ equals the Top-$K$ value of the standard MoE baseline (\eg, $K_s=1$ and $K_g=1$ for a Top-$2$ routing setup), ensuring the performance gains are not driven by an inflated computational budget, but rather by role-aware capacity allocation.

\subsection{Prototype-Guided Semantic Routing}

We introduce a set of learnable prototypes $\mathcal{P}=\{\mathbf{p}_1,\ldots,\mathbf{p}_{M_s}\}$ that act as a semantic bottleneck between continuous VAE features and discrete semantic experts. Inspired by SRA2~\cite{wang2026vae}, we use clean VAE features as a cost-free guidance space for the rich visual priors without relying on external representation models. Each prototype $\mathbf{p}_m$ serves as a semantic anchor for one semantic expert in the VAE feature space. To obtain a stable prototype topology, we decouple router learning from prototype optimization through two objectives.

\paragraph{Router alignment ($\mathcal{L}_{\rm align}$).}
For the $i$-th visual token, we compute a teacher assignment $\mathbf{q}_{i}$ by comparing its corresponding clean VAE feature $\mathbf{f}_{i}$ with the semantic prototypes using temperature-scaled cosine similarity. The resulting $\mathbf{q}_{i}$ is treated as a fixed target. Meanwhile, the router takes the DiT visual token $\mathbf{x}_{i}$ as input and predicts a semantic routing distribution $\mathbf{r}_{i}$ from the logits. We align this router prediction with the VAE-prototype induced soft target using Kullback--Leibler divergence:
\begin{equation}
\mathcal{L}_{\rm align}=\frac{1}{N}\sum_{i=1}^{N}\operatorname{KL}(\mathbf{q}_{i}\,\|\,\mathbf{r}_{i}).
\end{equation}
This objective trains the router to assign visually similar tokens to consistent semantic prototypes, while the stop-gradient operation prevents the router objective from directly distorting the prototype topology.

\paragraph{Prototype optimization ($\mathcal{L}_{\rm proto}$).}
The prototypes are explicitly optimized to track the underlying visual feature manifold. We formulate this process by treating prototypes as particles on a hypersphere governed by two complementary forces, attraction (pull) and repulsion (push)~\cite{wang2020understanding}:
\begin{equation}
\resizebox{0.9\hsize}{!}{%
$
\mathcal{L}_{\rm proto} =
\underbrace{
-\frac{1}{N}\sum_{i=1}^{N}\tau_c\log\sum_{m=1}^{M_s}e^{\frac{\cos(\mathbf{f}_{i},\mathbf{p}_m)}{\tau_c}}
+
\frac{1}{M_s}\sum_{m=1}^{M_s}\left(1-\max_{\bar{\mathbf{f}}\in\mathcal{B}}\cos(\mathbf{p}_m,\bar{\mathbf{f}})\right)
}_{\text{Pull: keep prototypes on current and historical VAE features}}
+
\underbrace{
\frac{1}{M_s(M_s-1)}\sum_{j\neq k}\max(0,\cos(\mathbf{p}_j,\mathbf{p}_k)-\delta)
}_{\text{Push: separate prototypes to avoid redundancy}}
$%
}
\end{equation}
Specifically, the \textbf{pull} term attracts prototypes toward both current mini-batch features and recent historical features stored in the circular token bank $\mathcal{B}$~\cite{he2020momentum}, ensuring that prototypes remain close to meaningful regions of the VAE feature manifold and reducing dead modes. Conversely, the \textbf{push} term imposes inter-prototype repulsion with margin $\delta$, preventing multiple prototypes from collapsing onto redundant background-dominated regions. Together, these two terms encourage prototypes to form separated and well-covered centers over the visual feature distribution. While ProMoE~\cite{wei2025routing} also introduces prototype-based routing, we use prototypes as a bridge between DiT router logits and reconstructive VAE features, rather than relying on direct prototypical matching within DiT layers alone.

Conceptually, $\mathcal{L}_{\rm align}$ aligns the router predictions with VAE-induced semantic assignments, while $\mathcal{L}_{\rm proto}$ learns a stable and diverse prototype topology.

\paragraph{Group-aware load balancing.}
\label{sec:loss_free}

Following DeepSeek-V3~\cite{liu2024deepseek}, we employ a loss-free load-balancing strategy for the generic experts.
Importantly, semantic experts avoid explicit load balancing, instead achieving adaptive balance naturally through semantic alignment.

\subsection{Overall Objective}
The full training objective is formulated as $\mathcal{L}
    =
    \mathcal{L}_{\rm flow}
    +
     \lambda_{\rm align}\mathcal{L}_{\rm align}+\lambda_{\rm proto}\mathcal{L}_{\rm proto}$, where $\mathcal{L}_{\rm flow}$ denotes flow matching loss.
Notably, the proposed group-aware load balancing introduces no auxiliary optimization loss. The mechanism exclusively updates the non-gradient routing biases for the expert selection process.

\section{Experiments}
\label{sce:experiments}
\subsection{Implementation Details}
\label{sce:implementation_details}

We upcycle the low-noise checkpoint of Wan~2.2~\cite{wan2025} by replacing the dense feed-forward networks at even-numbered layers from 15 to 35 with sparse MoE layers, while keeping all other components unchanged. This alternating design balances stable global feature extraction with expert specialization~\cite{liu2025efficient}. Each MoE layer contains one continuously active shared expert and 100 routed experts, partitioned into 20 semantic and 80 generic experts. For each token, a Top-$8$ router activates $K_s{=}2$ semantic experts and $K_g{=}6$ generic experts. By matching the active intermediate dimension of the dense baseline, our model activates 14B of its 27B total parameters per forward pass, maintaining computational parity with the dense model.
We initialize experts from the dense FFN to preserve pre-trained generative priors and train all models under the same 80k-step optimization setting.

\subsection{Experimental Setup}
\label{sec:experiment_setup}

\paragraph{Same-source baseline setup.}
To isolate component contributions, we compare same-source variants sharing the Wan~2.2 low-noise checkpoint, training data, step count, and inference settings. Unless specified, all MoE variants share the same configuration as in Sec.~\ref{sce:implementation_details}. \textit{Dense Wan2.2-FT} finetunes the original dense model identically.
\textit{Wan2.2-MoE }converts Wan~2.2 into a standard MoE (100 generic experts, Top-$8$ routing), representing our model without the semantic branch. \textit{SplitMoE w/o ProtoGuidance (PG)} retains the 20/80 semantic-generic partition and dual-track routing but omits VAE teacher features, semantic prototypes, $\mathcal{L}_{\rm align}$, and $\mathcal{L}_{\rm proto}$. \textit{SplitMoE w/o Pull} and \textit{SplitMoE w/o Push} respectively remove the attractive and repulsive terms of $\mathcal{L}_{\rm proto}$. \textit{Full SplitMoE} incorporates all proposed components. This protocol effectively disentangles dense fine-tuning, sparse scaling, role-aware expert partitioning, and prototype-guided semantic specialization.

\paragraph{Baseline methods and evaluation metrics.}
We compare \method against a wide range of SOTA text-to-video generation models, including CogVideoX-1.5~\cite{yang2024cogvideox}, Mochi~\cite{genmo2024mochi}, HunyuanVideo~\cite{kong2024hunyuanvideo}, LongCat-Video~\cite{team2025longcat}, LTX-2~\cite{hacohen2026ltx}, Wan2.2~\cite{wan2025} and OmniWeaving (think)~\cite{pan2026omniweaving}.
All videos are generated using the default inference settings of each respective model with 81 frames.

We evaluate \method on two complementary benchmarks. VBench-2.0~\cite{zheng2025vbench}, which is an upgraded version of VBench~\cite{huang2024vbench}, assesses intrinsic faithfulness across five capability dimensions: Creativity, Commonsense, Controllability, Human Fidelity, and Physics, using a combination of state-of-the-art VLMs/LLMs and specialist anomaly detectors. T2V-CompBench~\cite{sun2025t2v} targets compositional generation quality via MLLM-based, detection-based, and tracking-based metrics across three dimensions: Consistent Attribute, Interaction, and Numeracy.

\begin{figure}[tb]
  \centering
  \includegraphics[width=\linewidth]{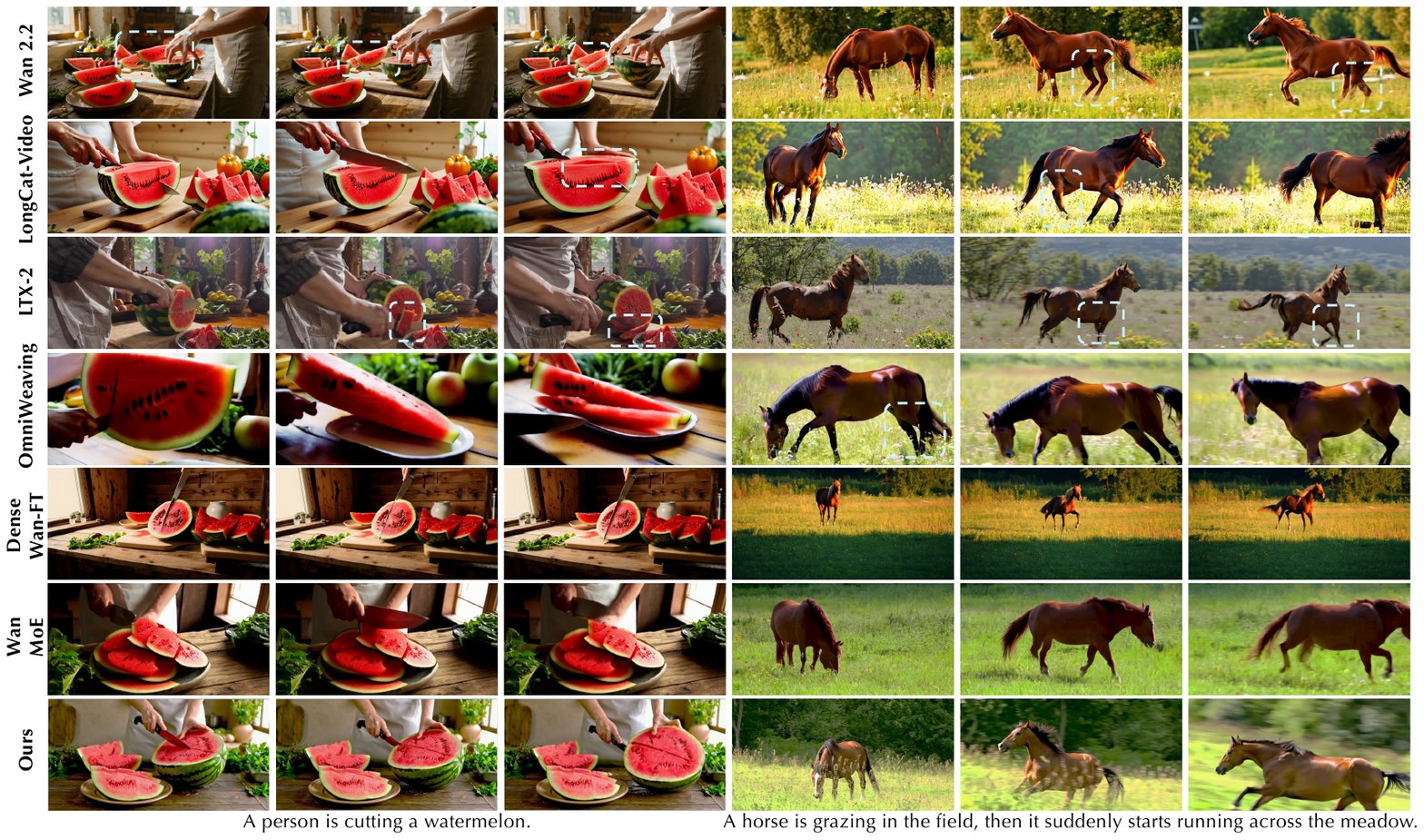}
  \vspace{-5em}
  \caption{Qualitative comparison with baseline methods. Zoom in for better comparison. We provide more examples and comparisons with more baseline methods in the video demo.}
  \vspace{-1.5em}
  \label{fig:qualitative_comparison}
\end{figure}

\subsection{Quantitative Comparison}
Table~\ref{tab:vbench2_results} reports quantitative results on VBench-2 and T2V-CompBench. Our method achieves competitive performance across multiple evaluation dimensions. Compared with Wan2.2, it improves Creativity and Human Fidelity, suggesting that the proposed routing design benefits diverse content generation and human-centric video synthesis. On T2V-CompBench, our method obtains competitive compositional-alignment results, indicating that the introduced MoE structure preserves text-video consistency under compositional prompts. VBench scores for LongCat-Video and HunyuanVideo follow the LongCat-Video paper, and T2V-CompBench scores for CogVideoX-1.5 and Mochi follow the T2V-CompBench benchmark.
Since independently developed models may differ in training data, scale, compute budget and post-training recipes, we use controlled same-source ablations to isolate the architectural contribution. The following section therefore focuses on MoE-specific comparisons.

\begin{table}
  \caption{Text-to-Video evaluation results on VBench-2 and T2V-CompBench, best scores in \textbf{bold} and second scores with \underline{underline}.
  A14B denotes MoE with 14B activated parameters.}
  \label{tab:vbench2_results}
  \centering
  \resizebox{\textwidth}{!}{
  \begin{tabular}{lccccccccc}
    \toprule
    & & \multicolumn{5}{c}{VBench-2} & \multicolumn{3}{c}{T2V-CompBench} \\
    \cmidrule(lr){3-7} \cmidrule(lr){8-10}
    Model name &\#Params.  & Creativity$\uparrow$ & \makecell{Common \\ Sense}$\uparrow$ &  \makecell{Control- \\ lability} $\uparrow$ &  \makecell{Human \\ Fidelity}$\uparrow$ & Physics$\uparrow$  & \makecell{Consist \\ attr.}$\uparrow$ & \makecell{Inter- \\ action}$\uparrow$ &\makecell{Nu- \\ meracy}$\uparrow$\\
    \midrule
    Dense Wan2.2-FT & 14B  & 49.57\% & 55.92\% & 30.37\% & 73.08\% & 54.66\% & 77.51\% & 61.08\%  & 36.71\%  \\
    Wan2.2-MoE &A14B  & 53.88\% & 57.15\% & 35.51\% & 73.49\% & 63.32\% & 78.82\%  & 63.89\% & 38.55\% \\ 
    \midrule
    SplitMoE w/o PG &A14B  & 51.05\% & 63.22\% & 34.83\% & 74.04\% & 55.89\% & 78.46\% & 61.96\%  & 38.57\%\\ 
    SplitMoE w/o Push  &A14B & 50.54\% & 62.75\% & 33.92\% & 73.60\% & 56.44\% & 77.29\% & 59.05\% & 37.15\%\\ 
    SplitMoE w/o Pull  &A14B & 52.23\% & 61.10\% & 35.67\% & 73.81\% & 56.83\% & 78.14\% & 61.15\% & 38.32\%\\ 
    \midrule
    CogVideoX-1.5 &5B  & 43.66\% & 58.19\% & 29.57\% & 72.14\% & 63.23\% &61.64\% &60.69\%  &37.06\%  \\
    Mochi &10B &39.84\% &61.08\%  &30.12\%  &78.91\% &57.61\% & 59.73\% &53.81\%  &27.18\%\\
    HunyuanVideo &13B  & 41.84\% & 63.44\% & 28.60\% & \underline{82.41\%} & 60.20\% &78.58\%  & 63.31\%  & 25.05\% \\
    LongCat-Video &13.6B  & \lowscore{54.73\%} & \textbf{70.94\%} & \lowscore{44.79\%} & 80.20\% & 59.92\% & 73.38\%    & 61.76\%     & \textbf{47.87\%}  \\
    LTX-2 &14B  & 50.39\% & \underline{67.09\%} & 38.09\% & 73.69\% & \textbf{76.71\%} & 69.93\%    & 47.98\%    & 25.83\%\\
    Wan2.2 &A14B  & \underline{58.35\%} & 62.45\% & \textbf{48.73\%} & 75.33\% & 69.07\% & \lowscore{83.19\%}   & \underline{73.29\%}     & 16.96\%  \\
    OmniWeaving &8.3B+7B  & 50.12\% & 61.89\% & 40.21\% & \lowscore{81.15\%} & \lowscore{67.21\%} & \underline{83.98\%}     & \textbf{73.94\%}      & 37.61\%\\ 
    \midrule
    Full \method (Ours) &A14B  & \textbf{58.46\%} & \lowscore{64.89\%} & \underline{45.72\%} & \textbf{84.47\%} & \underline{69.41\%} & \textbf{84.63\%}   & \lowscore{69.98\%}  & \underline{44.01\%}\\
    \bottomrule
  \end{tabular}
  }
\end{table}

\subsection{Qualitative Comparisons}
For a fair comparison, we use examples from VBench2. In Fig.~\ref{fig:qualitative_comparison}, most baselines follow text prompts well, with Wan2.2 showing high visual quality and LTX-2 producing realistic effects. In the watermelon-cutting case, however, baselines struggle with complex object interactions: Wan2.2, LongCat-Video, and LTX-2 show local distortions as the action progresses, while the dashed boxes highlight unnatural deformation and fusion of hand morphology and watermelon geometry. OmniWeaving further suffers from abrupt content changes and inter-frame blurring. In contrast, our method maintains temporal consistency, stable physical boundaries, and structural details throughout the action. The horse case evaluates complex motion transitions from standing to running, where baselines produce multi-leg artifacts and structural anomalies under large spatiotemporal changes. Our method yields a smoother transition and preserves accurate physiological structure, benefiting from our split experts and semantic routing.

\begin{table}[t]
\centering
\caption{Speed comparison.}
\label{tab:speed}
\resizebox{0.6\textwidth}{!}{
    \begin{tabular}{lccc}
    \toprule
    Method & Ours & Baseline MoE & Dense Model \\
    \midrule
    Total Parameters & 27B & 27B & 14B \\
    Activated Parameters & 14B & 14B & 14B \\
    Training Speed & 6.84s/it & 6.69s/it & 5.66s/it \\
    Inference Speed & 6.08s/it & 6.12s/it & 5.76s/it \\
    \bottomrule
    \end{tabular}
}
\end{table}

\subsection{Ablation Studies}

\paragraph{Computational overhead analysis.}
Tab.~\ref{tab:speed} compares training and inference speed under an iso-activated-parameter budget of 14B. Both our method and the baseline MoE show a slight per-iteration latency overhead over the dense model, mainly due to routing operations and memory bandwidth bottlenecks. Compared with the baseline MoE, our architecture adds negligible overhead and maintains highly comparable speed.

\begin{figure}[tb]
  \centering
  \includegraphics[width=\linewidth]{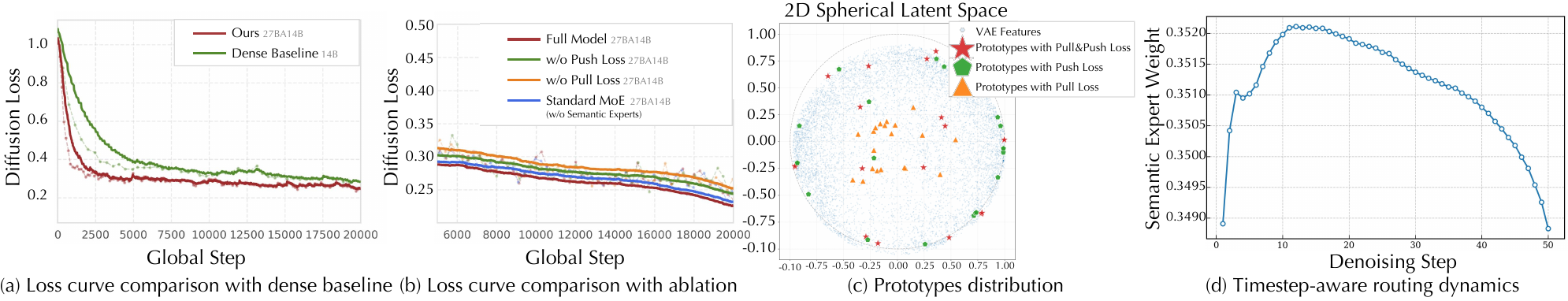}
  \vspace{-1.5em}
  \caption{Ablation study analysis.}
  \vspace{-10pt}
  \label{fig:ablation}
\end{figure}

\paragraph{Efficacy of MoE Upcycling.}
We evaluate sparse capacity scaling by comparing \method with a dense fine-tuning baseline. As shown in the first row of Table~\ref{tab:vbench2_results}, Dense Wan2.2-FT fine-tunes the low-noise Wan2.2 model on the same training data and evaluates it on VBench-2. Under the same active-parameter budget, the full \method outperforms this dense baseline across all evaluation dimensions, indicating that MoE upcycling raises capacity and the performance ceiling without increasing per-token activated parameters. The validation diffusion loss in Fig.~\ref{fig:ablation}(a) further supports this trend: \method converges faster, reaches comparable loss with roughly 70\% of the training steps, and maintains lower loss throughout training. These results show that our upcycling strategy preserves dense-model optimization stability while improving representational capacity through structure-aware sparse routing.
Notably, whereas the official Wan2.2 uses a high-/low-noise dual-model ensemble, Dense Wan2.2-FT is trained only from the low-noise checkpoint and thus naturally underperforms the full ensemble due to missing high-noise priors and reduced capacity.

\paragraph{Effect of structured semantic routing.}
We compare \method with two MoE baselines using similar expert configurations and parameter budgets. As shown in Table~\ref{tab:vbench2_results} and Fig.~\ref{fig:qualitative_comparison}, Wan2.2-MoE converts the low-noise Wan2.2 model into a standard MoE and fine-tunes it on the same data, while SplitMoE w/o PG keeps the same 20--80 expert partition as \method but removes prototype guidance. Their similar performance suggests that expert partitioning alone brings limited gains without explicit semantic guidance. In the left example in Fig.~\ref{fig:qualitative_comparison}, the watermelon struggles to maintain its complete semantics. In contrast, the full \method performs better on semantics-sensitive dimensions such as Human Fidelity and Physics, indicating that prototype-guided semantic routing improves expert specialization beyond naive sparse scaling. Fig.~\ref{fig:ablation}(b) supports this conclusion, as removing the semantic branch yields higher validation diffusion loss than the full model.

\begin{figure}[tb]
  \centering
  \includegraphics[width=1\linewidth]{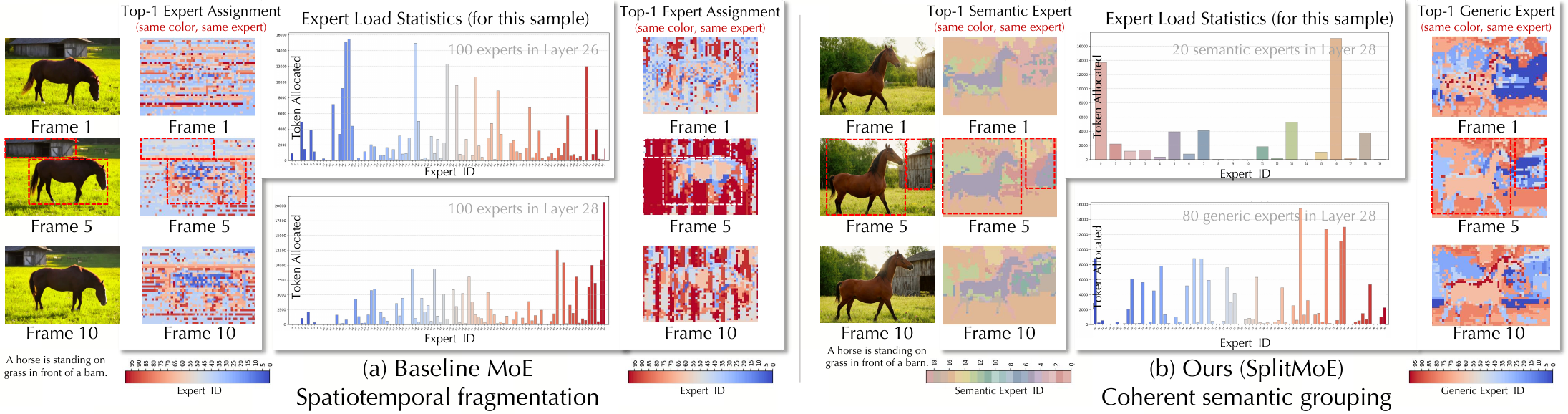}
  \vspace{-1.5em}
  \caption{Comparison of spatial routing and expert loads. (a) Baseline MoE scatters correlated patches across experts under strict capacity constraints, causing fragmented routing and poor spatiotemporal consistency. (b) \method routes coherent entities to Semantic Experts and residual textures to Generic Experts, achieving stable utilization without enforcing uniformity.
  }
  \vspace{-1.5em}
  \label{fig:routing_pathologies}
\end{figure}

\paragraph{Coupled effects of pull-push prototype guidance.}
We ablate the two complementary terms in $\mathcal{L}_{\rm proto}$ to test whether prototype guidance needs balanced attraction and repulsion. As shown in Table~\ref{tab:vbench2_results}, using only one term does not reliably improve quantitative metrics and can even underperform the variant without prototype guidance, since partial supervision may introduce a biased routing prior rather than a stable semantic partition. Fig.~\ref{fig:ablation}(b) shows the same trend in validation diffusion loss, where removing either term worsens optimization against the full model. The VAE-space visualization in Fig.~\ref{fig:ablation}(c) explains these failure modes. Without push, pull attracts prototypes toward dominant visual regions, preventing full coverage of the VAE manifold and limiting the semantic experts' representation space. Without pull, push drives many prototypes away from the data manifold, producing underused or dead semantic experts and weakening semantic routing. Thus, pull and push form a coupled mechanism: pull anchors prototypes to meaningful visual semantics, while push prevents redundant collapse. Together, they yield a balanced prototype layout that supports effective semantic expert specialization.

\paragraph{Routing pathologies in video MoE.}
Consistent with Fig.~\ref{fig:intro_motivation}, where standard MoE shows low semantic distinctiveness and intra-region routing dominance, Fig.~\ref{fig:routing_pathologies}(a) visualizes how this mismatch appears spatially and temporally. The Top-1 expert maps across frames show that coherent regions such as the horse and barn are not routed consistently. As highlighted by the dashed boxes, standard routing produces three artifacts: \textit{(i) Temporal jitter}, where similar regions switch experts across frames; \textit{(ii) Spatial striping}, where homogeneous areas are artificially split to satisfy load constraints; and \textit{(iii) Semantic fragmentation}, where adjacent tokens from the same object scatter across unrelated experts. These results indicate that rigid token-level balancing is ill-suited to video tokens. In contrast, Fig.~\ref{fig:routing_pathologies}(b) shows that \method assigns coherent object regions to stable semantic experts while using the generic branch to absorb residual textures and backgrounds, thereby reducing jitter, striping, and fragmentation.

\paragraph{Timestep-aware routing dynamics.}
We further track the routing weights assigned to semantic experts over a 50-step denoising trajectory in Fig.~\ref{fig:ablation}(d). The semantic weight peaks in the early high-noise stage, around steps 10--15, when the model establishes global layout, object placement, and coarse semantic structure. It then gradually decreases as denoising shifts toward local appearance and detail refinement. This coarse-to-fine pattern shows that the semantic branch is not redundant sparse capacity; instead, it learns a stage-dependent role, allocating semantic computation to the phases where high-level structure formation is most needed.

\section{Conclusion}
\label{sec:conclusion}
In this work, we identify the \textit{uniformity trap} as a key limitation of LLM-style MoE routing for video diffusion and propose \method, a split-role MoE that separates semantic abstraction from general visual residual modeling. With semantic-generic expert partitioning, prototype-guided routing, and pull-push regularization, \method improves convergence, routing coherence, and generation quality under matched activated-parameter budgets. Its timestep-aware routing reveals an emergent coarse-to-fine pattern, suggesting modality-aware sparse routing as a useful direction. Limitations include reliance on frozen visual features and sparse-routing overhead, motivating future work on adaptive partitioning, efficient distributed routing, and broader video generation tasks.

\clearpage

\bibliographystyle{plainnat}
\bibliography{references}

\end{document}